\documentclass{ar-1col}
\usepackage[numbers]{natbib}
\usepackage{url}
\usepackage{amsfonts}
\usepackage{gensymb}
\usepackage{float}
\usepackage{subfigure}
\usepackage{wrapfig}

\newcommand{\SE}{\mathrm{SE}}
\newcommand{\SO}{\mathrm{SO}}

\jname{Xxxx. Xxx. Xxx. Xxx.}
\jvol{AA}
\jyear{YYYY}
\doi{10.1146/((please add article doi))}

\begin{document}

\markboth{Robert Platt}{Grasp Learning}

\title{Grasp Learning: Models, Methods, and Performance}

\author{Robert Platt
\affil{Khoury College, Northeastern University, Boston, USA, 02120; email: rplatt@ccs.neu.edu}
}

\begin{abstract}
Grasp learning has become an exciting and important topic in robotics. Just a few years ago, the problem of grasping novel objects from unstructured piles of clutter was considered a serious research challenge. Now, it is a capability that is quickly becoming incorporated into industrial supply chain automation. How did that happen? What is the current state of the art in robotic grasp learning, what are the different methodological approaches, and what machine learning models are used? This review attempts to give an overview of the current state of the art of grasp learning research.
\end{abstract}

\begin{keywords}
grasp learning, grasping, machine learning, manipulation
\end{keywords}
\maketitle

\tableofcontents

\section{INTRODUCTION}

Machine learning has had a major impact on robotics over the last ten years. Nowhere is this more evident than in robotic grasping. Classically, grasping has been viewed as a planning or constrained optimization problem where the positions and forces of the robotic contacts are calculated in order to satisfy mechanical constraints such as force or form closure~\cite{murray1994mathematical,mason_2001}. However, with the advent of deep learning, it has become much more common to take a learning based approach where a neural network model infers a grasp strategy directly from image data without necessarily reasoning about object geometry or mechanics in a direct way. Approaches of this type have proven to be very effective when grasping in dense clutter or when the geometry of the objects is unknown.

The goal of this paper is to enable a researcher who is new to the field to quickly come up to speed with some key ideas in grasp learning. However, since this is a quickly evolving field, it is unclear what approaches are best in many cases of practical interest. Therefore, instead of presenting an overarching framework, this paper attempts to review and classify some of the major types of methods that have been pursued over the last five to ten years and to highlight a small number of papers that are representative of those categories. This review largely ignores papers older than five years in order to focus on contemporary methods. Moreover, it makes no attempt to cite or describe every single grasp learning paper and there are doubtless many important papers that have been missed. Nevertheless, this review will hopefully give the reader a broad perspective on a fast-moving field and a sense for where challenges may lie.

\section{GRASPING IN SE(2)}

A lot of work on grasp learning is in $\SE(2)$. Generally, a two fingered gripper is used and it is constrained to remain in a top-down configuration with the gripper fingers pointed straight down at the table. In this setting, the input is a top-down image of a scene (either RGB, RBGD, or depth) and the output is the $(x,y,\theta) \in \SE(2)$ coordinates of one or more hand poses from which a grasp is feasible. As such, the grasp learning problem can be expressed as the problem of finding a function $f : \mathbb{R}^{c \times h \times w} \rightarrow \SE(2)^k$ that maps from a $c$-channel $h \times w$ image onto a set of $k \geq 1$ feasible grasps.




\subsection{Sample and Test Methods}

\label{sect:sampletest}

This is one of the simplest approaches to grasp detection. Here, the focus is on classification of whether a grasp candidate is a good grasp or not. The input is an image $I$ (RGB, RGBD, or depth). Grasp candidates are often expressed in image coordinates $u,v,\theta$, where $u,v \in \mathbb{Z}^2$ denotes the grasp position in terms of its pixel coordinates and $\theta \in S^1$ denotes orientation of the gripper relative to the image plane. Given a grasp candidate, the neural network model predicts the grasp quality, i.e. whether the candidate is likely to be a good grasp or not. Sample and test methods work by sampling a large number of grasp candidates (e.g. exhaustively or using some heuristic) and then inferring the quality for each candidate. The algorithm returns the highest quality set of grasps.


\subsubsection{Lenz, Lee, and Saxena (2015)}

An early instantiation of the Sample and Test method was the work of Lenz, Lee, and Saxena~\cite{lenz_rss2013,lenz2015deep} where the authors use a small two layer fully connected neural network model for grasp classification. Grasps are represented as a \emph{grasp rectangle}, a rectangle in an image centered and oriented on the gripper in the grasp pose. This is analogous to an oriented bounding box that aligns with the grasp pose. The grasp rectangle is encoded to the neural network as a $24 \times 24$ image patch. A key challenge with this approach (and with the Sample and Test method in general) is the computational expense required to classify every possible grasp candidate in a scene, i.e. all possible grasp positions and orientations. This paper addresses the problem by using \emph{two} models: a lightweight model which is less accurate but cheap to evaluate and a heavyweight model which is more accurate but more computationally expensive. The lightweight model is evaluated on all candidates and only the top ranked ones are sent to the heavyweight model to be evaluated. The method was evaluated on the Cornell Dataset (Section~\ref{sect:cornell}) and found to reach a 75.6\% accuracy for an object-wise split, measured using an IOU threshold of 25\% and an orientation threshold of 30\% (see Section~\ref{sect:iou} for more details on these metrics). On a physical robot (a PR2~\cite{wyrobek2017origin}), they showed that the system could grasp household objects with an 89\% grasp success rate.

\subsubsection{Pinto and Gupta (2016)}
\label{sect:pinto}

Another variation on the Sample and Test method is the work of Pinto and Gupta~\cite{pinto2015supersizing}. Here, instead of a two-layer MLP, the authors adapted the AlexNet backbone~\cite{krizhevsky2012imagenet} (five convolutional layers and two fully connected layers) for grasp classification of image patches. As in Lenz et al.~\cite{lenz2015deep}, these image patches are centered on grasp candidates. However, whereas Lenz et al. classify \emph{oriented} image patches, Pinto and Gupta classify unrotated \emph{image-aligned} image patches. Specifically, the AlexNet backbone takes as input an unrotated image patch centered on a grasp candidate position and outputs set of 18 probabilities, denoting the probability that a grasp exists in each of 18 orientations (centered on the patch). The strength of this approach is that, unlike Lenz et al.~\cite{lenz2015deep} who must sample over both position and orientation, Pinto and Gupta only sample over grasp position, leading to a much smaller search space. This model was trained completely online with a physical robot which performed 50k grasp attempts over 700 hours of robot time. Ultimately, the authors demonstrated that the model could reach approximately 76\% classification accuracy on a held out dataset involving novel household objects and could grasp novel singulated objects with a 66\% grasp success rate.


\subsubsection{Mahler et al. (2017)}

The work of Mahler et al., dubbed the \emph{Grasp Quality CNN} (GQ-CNN) or simply \emph{DexNet 2.0}, is another variation on the Sample and Test method~\cite{mahler2017dex}. This work is similar to that of Lenz, Lee, and Saxena~\cite{lenz_rss2013} with the following differences. First, instead of using a two-layer MLP, Mahler et al. use a convolutional neural network with four convolutional layers and two fully connected layers. Like Lenz et al.~\cite{lenz2015deep}, this model takes a $32 \times 32$ oriented image patch as input, it produces a binary grasp quality prediction as output, and it is trained using a cross entropy loss. Second, instead of evaluating with a lightweight model and then a heavier model, Mahler et al. use only one model. However, rather than sampling exhaustively during grasp inference, they use the Cross Entropy Method (CEM)~\cite{rubinstein1997optimization} to locate high quality grasps. CEM iteratively performs rounds of grasp candidate sampling and evaluation using the model. After each round of evaluation, a small number of high quality grasps are selected and then further sampling is performed nearby those selected samples. The result is that DexNet 2.0 can effectively cover a large sample space with fewer samples than would otherwise be required. 
The authors evaluated DexNet 2.0 on a physical robot and obtained an 80\% grasp success rate on a set of ten novel household objects  and a 94\% grasp success rate on 40 objects drawn from the training set. 




\subsubsection{Summary}
The Sample and Test method is effective, but inference is computationally expensive because of the need to evaluate a large number of candidates.

\subsection{Regression to a Single Grasp}

This is another simple approach to grasp detection. The \emph{Regression to a Single Grasp} method takes a full image (not an image patch) as input and outputs a single prediction of the pose of a good grasp. The approach assumes that each scene contains at least one good grasp and formulates grasp prediction as a regression problem. A key limitation of this approach is that (by construction) it can detect only one grasp per scene.


\subsubsection{Redmon and Angelova (2015)}
\label{sect:redmon1}

Redmon and Angelova~\cite{redmon2015real} were the first to propose a grasp detection method of this type. Their model takes a single $225 \times 225$ image of a scene as input and outputs a vector that encodes the $(x,y,z,w,\theta)$ pose of a good grasp, where $x,y,z$ encode the 3D position of the grasp, $\theta$ encodes gripper orientation about the vertical axis, and $w$ describes gripper aperture (width). The model was trained using an L2 regression loss using the Cornell Dataset (see Section~\ref{sect:cornell}). Since Cornell contains multiple ground truth positives per image, the regression target for each image is chosen randomly from among these positives. Compared to Lenz et al.~\cite{lenz_rss2013}, this model must solve a more challenging inference problem because it must locate a single grasp in an entire image. In consequence, this model uses a full AlexNet~\cite{krizhevsky2012imagenet} pipeline rather than the two-layer model from~\cite{lenz_rss2013}. Nevertheless, the approach was shown to obtain an 84.9\% accuracy on a test set from the Cornell dataset using a 30 degree orientation threshold and a 25\% IOU threshold. The authors were able to improve upon this accuracy somewhat by incorporating object classification as an auxiliary loss and inferring a separate grasp for each of a set of regions (Section~\ref{sect:redmon2}).

\subsubsection{Kumra and Kanan (2017)}

Kumra and Kanan~\cite{kumra2017robotic} take an approach that is very similar to that of Angelova and Redmon~\cite{redmon2015real} except that they use a ResNet neural network backbone rather than AlexNet. The input is a single RGB image and the output is a single $x,y,z,w,\theta$ grasp prediction. The model is trained using an L2 loss relative to the nearest ground truth grasp in the training set. As would be expected, the ResNet backbone does better than AlexNet: Kumra and Kana obtained an 88.9\% accuracy on an object-wise split on the Cornell dataset using a 25\% IOU threshold.

\subsubsection{Summary}
Regression to a Single Grasp can be effective and efficient, but it is limited to detecting only one grasp per image.

\subsection{Region Proposal Methods}
\label{sect:sparsegrid}

A key idea in visual object detection was the Region Proposal Network (RPN), e.g. Faster R-CNN~\cite{ren2015faster} published in 2015. Shortly after that, several papers appeared that applied variations of this framework to grasp detection. The basic approach is to divide up the scene into a grid of cells and predict the location of exactly one grasp per cell. This can be viewed as a generalization of the Regression to a Single Grasp method -- here we solve a separate regression problem for each cell in the grid. 



\subsubsection{Redmon and Angelova (2015)}
\label{sect:redmon2}

In addition to the single grasp regression approach described in Section~\ref{sect:redmon1}, Redmon and Angelova \emph{also} applied the AlexNet model to the multi-grasp prediction problem~\cite{redmon2015real}. Here, the image is divided into a $7 \times 7$ grid. For each cell in the grid, we infer the probability that a grasp exists in that cell and the pose of that grasp relative to the center of the cell. While this was innovative work, a key drawback here was that the authors used a standard AlexNet model rather than a fully convolutional model. The fully connected layers at the end of AlexNet break the spatial structure present in the convolutional layers. This version of their method achieved 87.1\% accuracy on the Cornell dataset using a 25\% IOU threshold.


\subsubsection{Johns, Leutenegger, and Davison (2016)}

The approach proposed by Johns et al.~\cite{johns2016deep} is very similar to the multi-grasp approach of Redmon and Angelova. They use an AlexNet model to infer grasp orientation for each cell in a $45 \times 34$ grid. Orientation per cell is discretized into six categories and is trained using a binary cross entropy loss. Relative to Redmon and Angelova, the key distinctions here are: 1) the problem is cast as a set of binary classification problems over grasp position and orientation rather than an L2 regression problem; 2) the model does not infer offset relative to the center of each cell -- grasp positions are assumed always to be at the center of a cell. This was the first Region method to be tested on a physical robot and achieved an 80.3\% grasp success rate for singulated objects on a Kinova MICO arm.


\subsubsection{Zhou et al (2018)}

This method uses a fully convolutional ResNet-50 model~\cite{he2016deep} to infer grasp poses over a grid~\cite{zhou2018fully}. It outputs a coarse feature map that associates each cell in the grid with a binary prediction of whether a grasp exists in that cell and and prediction about the relative position and orientation of the grasp relative to the cell. It is trained using a combination of cross entropy and and L1 regression loss. 
Zhou et al. evaluate against the Cornell dataset, achieving 96.6\% accuracy for a 30 degree orientation threshold and a 25\% IOU threshold on held out object categories.

\subsubsection{Chu et al. (2018)}

Where as Zhou et al.~\cite{zhou2018fully} use a single ResNet-50 model, Chu et al.~\cite{chu2018real} use a sequence of two ResNet-50 models. The first generates region proposals and the second generates offsets. Each block has its own loss function: a combination of a cross entropy loss and an L1 regression loss. They evaluate their model on the Cornell dataset and obtain a 96.1\% detection accuracy on a per-object basis using RGBD input images (orientation threshold of 30 degrees and 25\% IOU threshold). Notice that this is slightly worse than Zhou et al.~\cite{zhou2018fully} on the same dataset and suggests that the greater model complexity of Chu et al.~\cite{chu2018real} does not help. The method was demonstrated to grasp singulated objects on a physical system with a 89\% success rate.


\subsubsection{Summary}
Region proposal methods naturally extend object detection methods to grasping. Equipped with a high capacity model like ResNet-50 (e.g. Zhou et al.~\cite{zhou2018fully}), these methods outperform both Sample and Test and Regression to a Single Grasp methods.


\subsection{Fully Convolutional Models}
\label{sect:fullyconv}

While some of the methods of Section~\ref{sect:sparsegrid} are fully convolutional, they only do inference over a sparse grid of positions in the workspace. In contrast, \emph{Fully Convolutional} grasping methods infer graspability at each pixel in the input image. In the simplest case, this amounts simply to adding convolution transpose layers to the end of the model, e.g. as in Morrison et al.~\cite{morrison2018closing}.


\subsubsection{Zeng et al (2018)} 

One of the first papers to explore this approach is the work of Zeng et al.~\cite{zeng2018robotic} who use a ResNet-101 backbone to infer planar grasp pose based on a single top-down RGBD image of a cluttered bin picking scene. The model is explored in the context of two end effectors: a suction effector and a gripper. In the case of the suction effector, the ResNet-101 outputs a heatmap over the scene that encodes grasp quality at a pixel level. The higher the output value for a particular pixel, the more likely it is that a good suction grasp can be performed there. Grasp inference for the gripper is more challenging because it requires specification of the gripper orientation about the approach direction. This is one of the most interesting aspects of the approach. Rather than inferring grasp quality for a set of different gripper orientations, Zeng et al. rotate the input image into 16 orientations which are passed to the ResNet model as a batch. These 16 scene orientations correspond to 16 gripper orientations. 
This is an example of \emph{canonicalizing} the input image -- transforming the input so that different grasp orientations are aligned into a single orientation about which the model can reason more easily. This system was an entry in the 2017 Amazon Robotics Challenge and scored first place in the Stowing Challenge with a 75\% pick success rate.




\subsubsection{Morrison et al (2018)} 
\label{sect:morrison1}

This model, dubbed \emph{GG-CNN}~\cite{morrison2018closing}, consists of three convolutional layers followed by three convolution transpose layers. It takes a $300 \times 300$ depth image as input and produces as output a stack of three $300 \times 300$ feature maps indicating grasp quality, gripper width, and grasp orientation for a grasp at that pixel. It is trained using the Cornell Dataset augmented with random translations, rotations, and zooms. At test time, after the model infers the graspability of each pixel but before selecting a grasp, the quality map is smoothed by applying Gaussian blur. This paper obtains images using a camera mounted to the wrist of the robotic arm and applies the model in both an open loop and closed loop (30 Hertz) setting. In both cases, the model was able to produce a grasp success rate in dense clutter of approximately 92\% on the Kinova Mico. 





\subsubsection{Asif et al (2018)}

This model is unusual because it does not infer grasps directly -- it is really a segmentation model for grasp affordances, i.e. it outputs a segmentation mask around the parts of an object that are graspable~\cite{asif2018graspnet}. After generating the segmentation mask, the model computes a mean, orientation, and width of each segmented region and outputs this information as grasp coordinates. The model as a series of ``Dilated Dense Fire'' (DDF) blocks that each incorporate a dialation layer, an expand layer, a batch norm, and some skip connections. The model is trained using a pixelwise cross entropy loss on ground truth segmentation masks. They integrate the DDF blocks into a U-Net like model, ultimately obtaining an 90.2\% grasp accuracy on the Cornell dataset measured with respect to a 25\% IOU threshold.


\subsubsection{Satish et al (2019)} 

In followup work to that of Mahler et al.~\cite{mahler2017dex}, the authors reformulated their approach using a fully convolutional model dubbed \emph{FC-GQ-CNN}~\cite{satish2019policy}. This model is comprised of a series of seven convolutional layers (no convolution transpose layers) and the output is a dense map of pixelwise probabilities of grasp success. Each pixel in the output is associated with a vector over gripper orientation that describes the probability of grasp success as a function of orientation. This model was trained using the ``network surgery'' method where fully connected layers are converted into $1 \times 1$ convolutional layers post-training. The neural model is actually a variant of the DexNet 2.0 model and it was trained using a cross entropy loss on the DexNet 2.0 dataset~\cite{mahler2016dex}. Then, it was converted to a fully convolutional model by converting the fully connected layers using ``surgery''. An important aspect of this work is the way the $z$ coordinate (height above the table) is handled. The input is a stack of $96 \times 96$ images where each image in the stack encodes a different gripper height by subtracting the corresponding height from each image in the stack. This enables the model to reason about different heights as if they were the same height. This is another instantiation of the canonicalization idea of~\cite{zeng2018robotic}. However, instead of rotating the input image, Satish et al. add a height offset. Ultimately, this model achieved an 87.5\% success rate grasping novel adversarial singulated objects and an 85.6\% success rate grasping novel objects in clutter.


\subsubsection{Kumra, Joshi, and Sahin (2020)}

Here, the authors propose a deep residual model (subbed \emph{GR-ConvNet}) that is similar to the fully convolutional models already described, but is larger (more layers and more parameters)~\cite{kumra2020antipodal}. Like~\cite{morrison2018closing}, the model outputs feature maps over grasp quality, orientation, and gripper aperture. However, this model incorporates several residual blocks sandwiched between an encoder which downsamples and a decoder which upsamples back to the original resolution. The model was evaluated on both the Cornell dataset and the Jaquard dataset using the IOU metric with a 25\% threshold. In both cases, they found their model outperformed the methods described earlier (when evaluated on Jaquard and Cornell), obtaining 94.6\% grasp classification accuracy on the Jacquard dataset (for RGBD images) and a 96.6\% accuracy on Cornell. Ultimately, they evaluated grasp success rates on a physical Baxter robot. They obtained a 95.4\% grasp success rate on singulated household objects, a 93\% grasp success rate on singulated adversarial objects, and a 93.5\% grasp success rate in clutter. 


\subsubsection{Zhu et al. (2022)}
\label{sect:zhu1}

Another fully convolutional grasp model that bears mentioning is the work of Zhu et al.~\cite{zhu2022grasp} who leverage an $\SO(2)$ fully convolutional equivariant neural model. This model is similar to the model of Satish et al.~\cite{satish2019policy} except that the model incorporates $\SO(2)$ equivariant convolutional layers that encode rotational symmetries directly into the neural network model. As a result of the additional rotational structure, this approach is dramatically more sample efficient than the other fully convolutional models described earlier. The authors demonstrated that the approach can learn a good grasp function with as few as 500 grasp demonstrations. As a result, the method can easily be used to learn to grasp directly on a physical robot (see Section~\ref{sect:zhu2}). This method was shown to be able to grasp ``easy'' objects in clutter with a 95\% success rate and ``hard'' objects in clutter with an 87\% success rate.

\subsubsection{Summary}

Currently, fully convolutional models are probably the best model choice for $\SE(2)$ grasp detection. A key idea is \emph{canonicalization}, an approach to inference over different grasp orientations and heights by modifying the input images so that different orientations or heights appear at a canonical orientation or height~\cite{zeng2018learning,satish2019policy}.

\subsection{Policy Learning for Grasping}
\label{sect:policylearning}

While most learning approaches to robotic grasping focus on identifying the final hand pose to be reached just prior to performing a grasp, there has also been work focused on policy learning for grasping. In this setting, the objective is to learn a closed loop policy that takes image input (and potentially proprioceptive and haptic input as well) and outputs effector velocities that should be taken in order to reach a grasp. Formally, this policy is a mapping $\pi : \mathbb{R}^{c \times h \times w} \rightarrow \dot{X} \times \dot{\Theta}$, where $\dot{X} \subseteq \mathbb{R}^3$ is the space of translational gripper velocities and $\dot{\Theta} \subseteq \mathbb{R}$ is the space of angular gripper velocities about the vertical axis. 


\subsubsection{Levine et al (2016)}
\label{sect:levine2016}

This work~\cite{levine2016learning,levine2018learningijrr} is essentially an application of Monte Carlo Reinforcement Learning (RL)~\cite{sutton_barto_book} to the grasping problem. State is a $472 \times 472$ RGB image of the scene. Action is a vector in $\dot{X} \times \dot{\Theta}$, the commanded Cartesian translational gripper displacement and an angular displacement about the vertical axis of the gripper (the gripper is constrained to point down along the vertical axis). The $Q$-function is modeled using a standard convolutional architecture (called the ``prediction function'' in the paper), which takes both state and action as input and outputs a scalar value. After executing a grasp episode (lasting at most 10 time steps), the $Q$-function is trained by pairing the state-action pairs experienced during the episode with the final outcome of the episode (either zero or one indicating grasp success or failure). Whereas most RL models are trained using an L2 loss, here the authors used a cross entropy loss, reflecting a view of the $Q$ function as a probability of grasp success rather than expected reward~\footnote{This view is only possible because of the sparse binary reward function used here.}. A key challenge here is estimating which action to take from a given state. This is hard because, unlike DQN~\cite{mnih2013playing}, action is an input to the model. This problem was addressed using the cross entropy method (CEM)~\cite{rubinstein1997optimization} where actions are iteratively sampled in order to find a maximal point. During grasping, the agent executed for at most 10 time steps at a rate of between 2 and 5 Hertz.

The authors of this paper made a couple of changes to the standard Monte Carlo RL formulation to improve sample efficiency. First, for state-action pairs belonging to episodes that terminated in a successful grasp, the actual action taken was ignored and replaced with the calculated displacement between the current gripper pose and the final gripper pose at the grasp. This reduces the credit assignment problem~\cite{sutton_barto_book} and makes the problem feel more like a contextual bandit. However, note that this trick is only possible when actions are transitive, as they are in this application, i.e. the net effect of a sequence of displacements is equal to the sum of the displacements. The second way the authors improved sample efficiency was to the use following two heuristics to control the gripper near the point of grasping. When the $Q$ function indicated that the probability of a successful grasp with zero gripper velocity was greater than 90\%, the first heuristic caused the robot to close the gripper. When the probability of a grasp success dropped to less than 50\%, the second heuristic caused the robot to open the gripper and raise it some distance above the table. One of the most interesting aspects of this system was that it was trained completely on physical experiences produced by a ``farm'' of approximately 14 robotic arms (this aspect of the work is detailed in Section~\ref{sect:levine2nd}). Ultimately, this system was able to produce grasp success rates of between 80\% and 90\% on physical robotic systems. Specifically, the first 10 objects picked from a 30 object bin in clutter were grasped with a 90\% success rate. The first 20 objects were grasped with a 82.5\% success rate and the total 30 objects in the 30 object bin were also grasped with a 82.5\% success rate.

\subsubsection{Kalashnikov et al. (2018)}

This method, dubbed \emph{QT-OPT}~\cite{kalashnikov2018qt}, is similar to that of~\cite{levine2016learning,levine2018learningijrr}, but it uses a standard TD learning RL approach rather than a Monte Carlo approach. The problem setup is nearly the same as that of Levine et al.~\cite{levine2016learning}: state is a $472 \times 472$ RGB image and action is a displacement $\dot{X} \times \dot{\Theta}$ coupled with an action that opens or closes the gripper. The model is also essentially the same as that used in~\cite{levine2016learning} -- a convolutional model that takes both state and action as input. As before, CEM is used to find actions that maximize the $Q$ function. As with~\cite{levine2016learning}, the model used in this paper was trained on physical robots over an extended period of time. However, \emph{unlike}~\cite{levine2016learning}, this work uses the standard TD target, $r + \gamma \max_{a \in A} Q(s,a)$~\cite{sutton_barto_book}. Another key difference relative to~\cite{levine2016learning} is that the gripper displacement action used on a given time step is \emph{not} replaced with the positional difference relative to the last time step. This enables this model to reason about the gripper \emph{trajectory} rather than just its end point. As a result of these differences relative to~\cite{levine2016learning}, this method is able to learn pregrasp manipulation strategies that can push objects as part of the grasp strategy in order to improve performance. The authors documented several pregrasp pushing interactions that the method learned to improve grasp outcomes. This method was evaluated in the same setting as~\cite{levine2016learning}. It achieved an 88.0\% grasp success rate for the first 10 objects picked from a 30 object bin, 88.0\% for the first 20 objects picked from the bin, and a 76\% success rate for all 30 objects grasped from the bin.

\subsubsection{Viereck et al. (2017)}

Unlike the methods of Levine et al.~\cite{levine2016learning} and Kalashnikov et al.~\cite{kalashnikov2018qt} which use reinforcement learning, the approach proposed by Viereck et al.~\cite{viereck2017learning} clones a $Q$ function produced by a planner. The method is trained in simulation. For each state-action pair, the $Q$ function infers the Euclidean distance between the gripper pose and the nearest good grasp pose after having executed the action. This model is trained by randomly generating grasp scenes and grasp poses from which the nearest good grasp is calculated by the planner. The model executes at approximately five Hertz and was demonstrated to be able to grasp objects that had been moved during grasp servoing. The method was demonstrated to be able to grasp a small set of novel household objects in clutter with an 88.9\% grasp success rate.


\subsubsection{Morrison et al. (2018)} 

Recall that the method of Morrison et al.~\cite{morrison2018closing} (see Section~\ref{sect:morrison1}) infers grasp pose pose, quality, and aperture using a fully convolutional model. This model can be used both in open-loop and closed-loop fashion. In closed-loop mode, the model is queried at approximately 30 Hertz and the poses of the three highest quality grasps are obtained. The method then executes a small displacement toward the nearest of these three grasps and the process repeats. Morrison et al. demonstrated that, like Viereck et al.~\cite{viereck_corl2017}, this method can be used to grasp moving objects. This method was demonstrated to reach an 87\% grasp success rate for the same cluttered grasp setting as in~\cite{viereck_corl2017}.



\subsubsection{Summary}

Although the policy learning problem is more challenging than the standard problem of just learning a grasp detection function, it is more powerful in at least two ways. First, a grasp policy has the ability adapt to object motion during grasping, i.e. to perform dynamic grasping. Second, a grasp policy has the potential to learn pregrasp interactions such as pushing or nudging in a way that could improve grasp outcomes.





\subsection{Approaches That Are Trained on a Physical Robot}

Whereas most grasp learning methods are trained in simulation, a few are trained completely on physical robotic systems. All of these methods are based on $\SE(2)$ grasping pipelines.

\subsubsection{Pinto and Gupta (2016)}

This approach~\cite{pinto2015supersizing} was described in Section~\ref{sect:pinto} and is one of the first grasp learning methods trained entirely on a physical robotic system. Training data was generated by executing pseudo-random grasp attempts on a Baxter robot. Objects were presented on a cluttered table top and segmented from the background. Training grasps were generated by sampling an object and then sampling a grasp pose around the object, with both samples taken uniformly at random. All together, approximate 50K grasp attempts were obtained this way over the course of 700 hours using household objects drawn at random from a total of 150 different physical objects.

\subsubsection{Levine et al. (2016)}
\label{sect:levine2nd}

This approach~\cite{levine2016learning,levine2018learningijrr}, which was described extensively in Section~\ref{sect:levine2016}, is essentially an application of Monte Carlo Reinforcement Learning~\cite{sutton_barto_book} to closed-loop grasp learning. However, one of the most interesting aspects of the work is the fact that it was trained completely using grasp data generated over the course of 800K grasp attempts (episodes) obtained by running between 6 and 14 robots nearly continuously over the course of two months. Approximately half this data was obtained by executing a random policy with small hand coded heuristics which produced a grasp success between 10\% and 30\% of the time. The remainder was collected by rolling out partially trained versions of the learned model. Ultimately, this system was able to produce grasp success rates of between 80\% and 90\% on physical robotic systems.

\subsubsection{Kalashnikov et al. (2018)}

Another approach where the model was trained using a large amount of physical robot experience was the QT-OPT method of Kalashnikov et al.~\cite{kalashnikov2018qt}. This method was trained over the course of 800 robot hours using 7 LBR IIWA robots over four months. Grasping was performed in cluttered scenes containing between four and ten objects. Data was obtained by initially rolling out a random policy and later rolling out partially trained versions of the learned policy.




\subsubsection{Zhu et al. (2022)}
\label{sect:zhu2}

The method of Zhu et al.~\cite{zhu2022grasp} (see Section~\ref{sect:zhu1}) obtains high sample efficiency by leveraging an equivariant neural network model. As a result, the authors were able to demonstrate that the method could learn to grasp on a single UR5 robot in just 500 grasps of Boltzmann exploration, obtained over the course of approximately an hour and a half. This is much faster than any of the other methods described above.




\subsection{Summary of $\SE(2)$ Methods}

Table~\ref{table:se2_comparison} summarizes the performance of the various methods described in this section in terms of performance on the Cornell detection benchmark and grasp success rates on a real robot. \emph{Cornell IOU} is the Intersection Over Union performance (see Section~\ref{sect:iou}) of the method for an object-wise training/test split and a 25\% IOU threshold, 30\% orientation threshold. \emph{Singulated Objects GSR} is the grasp success rate on a physical robot for novel objects (not seen during training) presented either in isolation or sufficiently far apart to allow for easy segmentation. \emph{Clutter Pile GSR} is the grasp success rate on a physical robot for novel objects presented in clutter. The five vertical groupings of the methods correspond to the methods covered in Sections~\ref{sect:sampletest} through~\ref{sect:policylearning}, respectively.

\begin{table}[h]
\tabcolsep7.5pt
\caption{Comparison between $\SE(2)$ grasp methods.}
\label{table:se2_comparison}
\begin{center}
\begin{tabular}{@{}l|c|c|c|c@{}}
\hline
&  & Singulated  & Clutter &   \\
& Cornell & Objects  & Pile & Object   \\
Method & IOU$^{\rm a}$  & GSR$^{\rm b}$ & GSR$^{\rm c}$ & Robot \\
\hline
Lenz, et al.~\cite{lenz2015deep} & 75.6\% & 89\% &  &  PR2 \\
Pinto and Gupta~\cite{pinto2015supersizing} &  & 66\% &  &  Baxter \\
Mahler et al.~\cite{mahler2017dex} &  & 94\% / 80\%$^{\rm d}$  & &  ABB Yumi \\
\hline
Redmon and Angelova~\cite{redmon2015real} & 87.1\% &  &   &  \\
Kumra and Kanan~\cite{kumra2017robotic} & 88.9\% &  &     \\
\hline
Johns et al.~\cite{johns2016deep} & & 80.3\% &   & Kinova  \\
Zhou et al.~\cite{zhou2018fully} & 96.6\% &  &   &  \\
Chu et al.~\cite{chu2018real} & 96.1\% & 89\% &   & Custom \\
\hline
Zeng et al.~\cite{zeng2018robotic} &  &  &  75\%$^{\rm e}$ & ABB IRB\\
Morrison et al.~\cite{morrison2018closing} &  & 92\% & 87\%  & Kinova  \\
Asif et al.~\cite{asif2018graspnet} & 90.2\% & &   &  \\
Satish et al.~\cite{satish2019policy} & &  & 85.6\% & UBB Yumi \\
Kumra et al.~\cite{kumra2020antipodal} & 96.6\% & & 93.5\% & Baxter \\
Zhu et al.~\cite{zhu2022grasp} & & & 87\% & UR5 \\
\hline
Levine et al.~\cite{levine2016learning} & & & 82.5\%$^{\rm f}$ & Custom \\
Kalashnikov et al.~\cite{kalashnikov2018qt} & & & 88.5\%$^{\rm f}$ & Custom \\
Viereck et al.~\cite{viereck2017learning} & & & 88.9\% & UR5 \\
\hline
\end{tabular}
\end{center}
\begin{tabnote}
$^{\rm a}$Cornell IOU as reported here is measured using a 25\% IOU threshold and a 30\% orientation threshold.\\
$^{\rm b}$Grasp success rate for singulated novel objects not seen during training.\\
$^{\rm c}$Grasp success rate for objects piled in clutter, as in the bin picking setting.\\
$^{\rm d}$80\% GSR for novel objects; 94\% GSR for objects seen during training.\\
$^{\rm e}$Pick success rate for the 2017 Amazon Robotics Challenge (ARC) final stowing challenge.\\
$^{\rm f}$GSR for first 20 objects removed from a 30 object bin.
\end{tabnote}
\end{table}


The only column in Table~\ref{table:se2_comparison} that should be used to compare methods in a precise way is \emph{Cornell IOU}. All methods that use this benchmark are using exactly the same test set. It is critical to recognize that the same is \emph{not} true for the grasp success rates we report in \emph{Singulated Objects GSR} and \emph{Clutter Pile GSR} in Table~\ref{table:se2_comparison}. Those are the results from physical robotic experiments and depend on the object set used for evaluation and the physical robotic setup. A paper that evaluates grasp success using objects that are challenging to grasp is disadvantaged relative to a paper that uses easy objects. Different robotic setups can also have an effect. While these differences have even more pronounced effects in the $\SE(3)$ (Table~\ref{table:se3_comparison}), they are also important here. 

\subsubsection{Discussion}

Based on Table~\ref{table:se2_comparison}, the model used by Kumra et al.~\cite{kumra2020antipodal} stands out because it scores highly on Cornell and also has a high grasp success rate in clutter. However, one should be cautious here. The clutter grasp setting used in \cite{kumra2020antipodal} is probably easier than some others, for example Kalashnikov et al.~\cite{kalashnikov2018qt} or Satish et al.~\cite{satish2019policy}. Perhaps the main conclusion to draw is that the fully convolutional model class (the fourth vertical grouping in Table~\ref{table:se2_comparison}) is currently probably the most competitive choice for a $\SE(2)$ grasping model architecture. Three of the other models -- sample and test, regression, and region proposal -- all seem somewhat outmoded now. However, policy learning is probably an area that will receive continued attention into the future because of its ability to learn closed loop policies rather than just to detect grasp poses.

\section{GRASPING IN SE(3)}

Given a point cloud or other volumetric representation of a scene, the problem of grasping in $\SE(3)$ is to identify a set of $\SE(3)$ hand poses from which a grasp is feasible. As such, the $\SE(3)$ grasp learning can be expressed as the problem of finding a function $f : \mathbb{R}^{3 \times n} \rightarrow \SE(3)^k$ that maps from a cloud of $n$ points in $\mathbb{R}^3$ to a set of $k$ grasp poses. Grasp learning in $\SE(3)$ is often approached using very different neural network models than it is in $\SE(2)$. Whereas fully convolutional are typically the best performing $\SE(2)$ grasp models, $\SE(3)$ grasp detection has been tackled using a variety of models including point models, 3D convolutions, graph neural networks, and variational autoencoders.

\subsection{Hypothesize and Test Methods}
\label{sect:hypandtestse3}


As was the case in $\SE(2)$ grasp learning, the earliest approaches to $\SE(3)$ grasp detection were hypothesize and test methods where grasp detection is decomposed into two phases. In the first phase, a large set of potential grasp candidates (6-DOF hand poses that could be good grasps) are generated. In the second phase, a classification model is queried which infers whether each candidate is a good grasp. This model is queried separately (or in batches) over the candidates.


\subsubsection{Gualtieri et al. (2016)}

One of the earliest approaches to novel object grasping in six dimensions is the work of Gualtieri et al. who proposed a method dubbed \emph{Grasp Pose Detection} (GPD)~\cite{gualtieri2016high,tenpas_ijrr2017}. This is a hypothesize and test method where grasp candidates are generated by sampling $\SE(3)$ hand poses in the vicinity of points in the cloud. In order to sample a grasp, this approach first samples a point from the cloud and then estimates the local Darboux coordinate frame at the sample~\footnote{The Darboux reference frame at a point on a smoothly curved surface is defined by the orthonormal basis consisting of the surface normal, the axis of principle curvature, and the binormal direction.}. A grasp candidate is then sampled using heuristics defined with respect to the Darboux frame. Candidates where the gripper is in collision with points in the cloud or do not contain a point from the cloud within the closing region of the gripper are pruned. Once sampled, the system makes a binary prediction about whether each candidate is a grasp or not. This prediction is based on a representation of the volumetric region contained between the fingers of the gripper, as expressed in the reference frame of the grasp candidate. This volume is expressed as a stack of three orthographic projections and is input to a small convolutional model. This gripper-frame representation of the grasp can be viewed as another type of canonicalization where grasps in different poses are expressed in a constant reference frame. Under ideal conditions, this approach was shown to have approximately a 93\% grasp success rate in dense clutter for novel household objects. 


\subsubsection{Liang et al. (2019)}

This method, dubbed \emph{PointNetGPD}~\cite{liang2019pointnetgpd} is very similar to GPD. The key distinction is the grasp classification model. Both PointNetGPD and GPD made grasp predictions based on a representation of the part of the cloud contained in the volume between the gripper fingers. However, whereas GPD represents this volume using an ordinary convolutional model operating on orthographic projections of the contained points, PointNetGPD uses a PointNet model. The method was compared with GPD and shown to have somewhat higher grasp success rates. However, like GPD, it requires a separate evaluation of the model for each batch of grasp candidates.


\subsubsection{Mousavian, Eppner, and Fox (2019)}

Another Hypothesize and Test method is the work of Mousavian, Eppner, and Fox~\cite{mousavian20196}, dubbed \emph{GraspNet}. In contrast to GPD and PointNetGPD which generate grasp samples using heuristics, this work uses a variational autoencoder (a VAE~\cite{kingma2013auto}) to generate the grasp candidates. The method assumes that it is given a segmented partial point cloud that corresponds to the object of interest. The VAE model is defined as follows. Let $G$ denote a grasp, $X$ denote a point cloud, and $z$ denote a latent variable. Both encoder and decoder are conditioned on the point cloud. The encoder is a model $Q(z | X,g)$ and the decoder is a model $P(G | X,z)$. This model is trained using the standard VAE loss that combines a reconstruction term with the VAE regularizer which biases the latent variable toward the standard Normal distribution. Both the encoder and decoder are implemented using PointNet++~\cite{qi2017pointnet++}. The gripper pose is encoded to the model as a set of points on the surface of the gripper that are added to the point cloud, with special features indicating their identity as gripper points. After the model is trained, grasp candidates can be generated by sampling from the standard Normal distribution and evaluating the decoder $P$. After sampling a set of grasp candidates from the VAE decoder, they are evaluated using a PointNet classifier. Again, points belonging to the gripper are added to the object point cloud. Then, the model is passed to a PointNet model which infers a binary grasp quality (trained with a cross entropy loss). Like PointNetGPD, a separate pass through the PointNet model is required for each sample. After identifying a set of good grasps, this method lastly performs a refinement step that makes small adjustments to grasp pose. 
The method is evaluated on four categories of household objects: boxes, cylinders, bowls, and mugs where they obtained an 88\% grasp success rate in physical robotic experiments. The most important aspect of this work is its novelty. The idea of using a VAE to sample grasps is interesting and potentially powerful. Unfortunately, this is a heavyweight model and the need to evaluate and refine a large number of grasp samples can be computationally intensive.

\subsubsection{Murali, Mousavian, Eppner, Paxton, and Fox (2020)}

This is an extension of~\cite{mousavian20196} where the authors augment the grasp quality evaluator with a collision detector that is applied after a set of high quality grasps are identified~\cite{murali20206}. The collision detection model is similar to the grasp evaluation model, but the PointNet++ model takes as input a slightly large point region around the target object in order to model possible collisions with those objects. The model is shown to outperform~\cite{mousavian20196}. 

\subsubsection{Summary}
While effective, Hypothesize and Test methods can be computationally slow because they must evaluate a classification model for a large number of grasp candidates.




\subsection{Grasp Inference Using Point Models}

An alternative to the Hypothesize and Test method is to leverage a point-based model like PointNet~\cite{qi2017pointnet} or PointNet++~\cite{qi2017pointnet++} to infer grasp pose. These models take a point cloud of an object or an entire scene as input and infer a set of grasp poses in a single forward pass of the model. This avoids the need to repeatedly evaluate a grasp quality model separately for each grasp candidate.


\subsubsection{Qin et al. (2020)}

The work of Qin et al. is an unadorned instantiation of this idea, resulting in a model dubbed \emph{S4G}~\cite{qin2020s4g}. An unsegmented point cloud is passed into a PointNet++ model, producing a feature vector for each point. Then, a sparse point set is created by pooling feature vectors based on 3D distance. Finally, a prediction about grasp quality and orientation is made at each of the sparse points. The model infers a gripper orientation which is represented as a pair of vectors in $\mathbb{R}^3$ and converted into a rotation matrix using Gram Schmidt orthogonalization. A set of high scoring grasps are selected and the corresponding grasp poses identified. The method is found to outperform GPD and PointNetGPD in simulated experiments. Ultimately, they show a 77\% grasp success rate in dense clutter in a tabletop scenario using a Kinova Jaco2 arm.

\subsubsection{Sundermeyer et al. (2021)}

This model, dubbed \emph{Contact-GraspNet}~\cite{sundermeyer2021contact}, is similar to that of Qin et al.~\cite{qin2020s4g}. The authors use a PointNet++ model which takes a point cloud of the scene as input and infers grasp quality and orientation at a set of points. As in Qin et al., orientation is represented as a pair a vectors in $\mathbb{R}^3$ which are used to reconstruct a rotation matrix using Gram Schmidt orthogonalization. Probably the key difference relative to Qin et al. is that they constrain grasps to contact the object surface at the observed point rather than inferring simply whether a grasp exists anywhere nearby. The resulting model is shown to perform favorably relative to some of the authors' prior work~\cite{mousavian20196} with an 84.31\% grasp success rate for household objects presented in clutter.


\subsubsection{Wu et al. (2020)}

This model~\cite{wu2020grasp} also consists primarily of a single PointNet++ backbone also, but is a bit more complex than that of Qin et al.~\cite{qin2020s4g}. It takes a segmented partial point cloud of an object as input which is passed to the PointNet++ model which generates per-point features. Grasp candidates are generated based on the cross product between points in the cloud and the 3D grid of voxel positions in the workspace. Every pair of a cloud point with a voxel position is a potential grasp candidate. These candidates are pruned using geometric conditions. For each candidate, a score and regressed pose is generated. The method was compared primarily to the work of Mousavian et al.~\cite{mousavian20196} in terms of the success-rate\@k and coverage-rate\@k metrics (Section~\ref{sect:noutofm}). They achieve an 85\% grasp success rate for singulated household objects.

\subsubsection{Zhao et al. (2021)}

These authors extend the S4G model of Qin et al.~\cite{qin2020s4g}, producing a three stage model dubbed \emph{RegNet}~\cite{zhao2021regnet}. The first stage (the ``score network'') is essentially the same as S4G: a PointNet++ model is used to infer grasp quality at each point in the cloud. Then, for points determined to have a high grasp quality, the ``grasp region network'' infers grasp orientation as a categorical distribution over a set of approach directions and approach angles. This prediction is made based on a feature representation obtained by pooling the PointNet++ features over a neighborhood of points around the query point. Finally, in the third stage (``grasp refinement''), a canonical hand-centric crop is taken around high quality grasp points and used to infer a new quality prediction and a residual pose relative to the discrete pose found in the second phase. This is a complex model, but the three phases make sense. Their results show that, with a 79.34\% grasp success rate for challenging household objects presented in clutter, the model outperforms S4G~\cite{qin2020s4g} as well as PointNetGPD~\cite{liang2019pointnetgpd} and GPD~\cite{tenpas_ijrr2017}.



\subsubsection{Wei et al. (2021)}

This model~\cite{wei2021gpr} is similar to RegNet~\cite{zhao2021regnet} except that it has two stages instead of three. The first infers a set of grasp candidates and the second evaluates those candidates in a canonical hand-centric reference frame. The first stage takes a point cloud input of an entire scene as input and passes it to a PointNet++ backbone which infers the grasp quality of each point and the associated grasp orientation. Then, after doing non-maximum supression on detected grasps, each candidate is expressed as a cropped point cloud in a canonical hand-centric reference frame and classified using another (smaller) PointNet++ model. This model is shown to achieve a 69.2\% grasp success rate in clutter comprised of challenging household objects. It performs favorably relative to GPD~\cite{tenpas_ijrr2017} and PointNetGPD~\cite{liang2019pointnetgpd}, but is not compared against S4G~\cite{qin2020s4g}.

\subsubsection{Summary}
Grasp inference using Point based models like PointNet++ could have a bright future in $\SE(3)$ grasp learning since they have the ability to reason about grasps over an entire object or scene efficiently in a single pass. Currently, this work has been restricted to PointNet++ and related models, but a variety of other point based models exist that have not yet been explored in the context of grasping, e.g.~\cite{chen2021equivariant}.



\subsection{Grasp Inference Using 3D Convolutional Models}

Since fully convolutional models are state of the art in $\SE(2)$ grasp learning, it makes sense to try to extend the idea to $\SE(3)$ grasping. 3D convolutional models do exactly this. The input to the model is a voxelized occupancy grid or a voxelized truncated signed distance function (a TSDF). The output is a per-voxel estimate of the quality and pose of one grasp per voxel. Unfortunately, the 3D convolutions used by these models is typically computationally expensive. 


\subsubsection{Breyer et al. (2021)}

This approach, dubbed Volumetric Grasp Network (VGN) is analogous to the fully convolutional approaches described in Section~\ref{sect:fullyconv} where the 2D convolutional layers have been replaced with 3D convolutional layers~\cite{breyer2021volumetric}. The input is a truncated signed distance function (a TSDF): a $40 \times 40 \times 40$ voxel grid where each voxel is labeled with its distance to the nearest object surface (distance above a threshold are capped). Unlike~\cite{mousavian20196}, this model does not require segmented objects at the input -- it can handle complete scenes. The model consists of three 3D strided convolutional layers followed by three dense 3D convolutional layers interleaved with $2\times$ bi-linear upsampling. The results is a fully convolutional 3D model with three heads that predict grasp quality, orientation (represented as a quaternion), and gripper aperture. The loss function is a combination of a binary cross entropy loss and standard L2 regression. The model was evaluated on a Panda robot and was found to perform favorably relative to GPD~\cite{tenpas_ijrr2017}, reaching an 80\% grasp success rate for household objects presented in clutter. However, since their setup creates a point cloud by fusing points generated from six different viewpoints, these authors are evaluating their performance on an easier inference problem relative to many of the other works described here.


\subsubsection{Cai et al. (2022)}

Another method which relies heavily upon 3D convolutions is the method of Cai et al., dubbed the \emph{volume point network} (VPN)~\cite{cai2022real}. There are two stages of this method. In the first stage, a TSDF is input to a fully convolutional stack of 3D convolution layers which outputs a 3D map of equal resolution as the input. It is assumed that all grasps will approach the object along the estimated surface normal at the grasp point. Each voxel in the output is associated with a vector of quality estimates corresponding to a discrete set of approach orientations (measured about the surface normal). Once a set of good grasp poses are identified using the fully convolutional model, these grasps are refined using a second 3D fully convolutional model. These authors perform experiments in a cluttered grasping setting, demonstrating a 91\% grasp success rate for challenging household objects and a 78.43\% grasp success rate for 12 of the 13 Dex-Net adversarial objects.

\subsubsection{James et al. (2022)}

Because 3D convolutions are so computationally expensive, it is typically impractical to reason about the volume of the scene at a fine level of discretization, e.g. Breyer et al.~\cite{breyer2021volumetric} infer grasps at a resolution of just $40^3$. James et al. circumvent this problem using hierarchy~\cite{james2022coarse} in the context of $Q$ learning. Specifically, they represent the $Q$ function as a hierarchy of two $Q$ models which we denote as $Q_{coarse}$ and $Q_{fine}$. First, given a coarse voxelized representation of a scene, $Q_{coarse}$ is used to infer the approximate 3D position of a grasp $x_{coarse}$. Then, the volume in the neighborhood of $x_{coarse}$ is voxelized at a finer level of resolution and passed to $Q_{fine}$ which infers the exact position and orientation of the gripper needed to grasp. The key advantage of this approach is that it is possible to reason at a fine level of resolution without representing the entire scene this way. This this paper does not explicitly focus on grasping per se, there are no reported grasp success rates.

\subsubsection{Summary}

Grasp inference using 3D convolutions is attractive because is simple and analogous to successful approaches in $\SE(2)$. However, the computational burden of 3D convolution is a serious drawback.

\subsection{Grasp Detection via Shape Completion}

Grasping via shape completion is an intuitively attractive approach. If the 3D geometry of an object can be reconstructed from observations, then a variety of grasp planning methods can be used to find grasps~\cite{murray1994mathematical,mason_2001}. This approach has been explored by several researchers, but has not been as successful as other methods.

\subsubsection{Varley et al. (2017)}

The focus here is primarily on shape completion, rather than on the grasping problem~\cite{varley2017shape}. At the time this paper was published, shape completion was not well studied. The authors proposed a 3D convolutional model which did shape reconstruction on a $40^4$ voxel grid: three 3D convolutions followed by some fully connected layers. The model was trained using a voxel-wise binary cross entropy loss. After performing completion, the model was converted into a mesh using the marching cubes algorithm~\cite{lorensen1987marching} and then smoothed using various gap filling methods and by minimizing the Laplacian of the surface. Given the mesh model, grasp detection happened using standard grasp planners available via GRASPIT!~\cite{miller2004graspit}. These authors report a 93\% grasp success rate on singlulated household objects.

\subsubsection{Lundell, Verdoja, and Kyrki (2019)}

A key problem with the approach of Varley et al. is that is fails to deal with uncertainty in the shape completion. Completing an entire 3D shape based on a partial point cloud is naturally an uncertain process and it seems that this uncertainty should be taken into account during grasp synthesis. This is exactly the idea that Lundell, Verdoja, and Kyrki explored~\cite{lundell2019robust}. The idea here is to develop a method of sampling from a distribution of possible shape completions rather than simply assuming the most likely completion to be accurate. By sampling different ways of completing a given shape, these authors can rank a set of grasps based on how many different possible completions each grasp would be valid for. The finally accepted grasp would be the one that would succeed most often over the space of different shape completions. This idea can be viewed probabilistically as marginalizing the grasp ranking over the distribution of possible shape completions. The shape completion model is a 3D convolutional UNet model that has a Monte Carlo Dropout at the center, which is used to sample different completions. In their experiments, the authors here reported a small improvement relative to Varley et al.~\cite{varley2017shape}, achieving a 58\% grasp success rate on singulated objects.

\subsubsection{Lundell, Verdoja, and Kyrki (2020)}

This paper uses scene completion to enable a 2D grasp detection method (in this case FC-GQ-CNN~\cite{satish2019policy}) to detect grasps in 6-DOF~\cite{lundell2020beyond}. This is a simple idea. First, they use scene completion to create a relatively complete model of the scene. Then, they render a height map of the scene from different possible grasp approach directions. Finally, they detect grasps in these different height maps using FC-GQ-CNN, thereby effectively detecting 6-DOF grasps. These authors do not report grasp success rates on a real robot.


\subsubsection{Yang et al. (2021)}

This paper uses a shape completion approach to improve grasps originally found using a standard grasp detection method~\cite{yang2021robotic}. First, a grasp candidate is detected using standard convolutional model that takes a stack of top-down images as input and outputs a single 6-DOF grasp detection (no shape completion information is used as this stage). This detection is refined by centering the grasp on the reconstructed points. The authors report a 84\% grasp success rate for singulated household objects presented on a table top.

\subsubsection{Van der Merwe, et al. (2020) and Jiang et al. (2021)}
\label{sect:merwe1}

It is not completely clear that the right application of shape completion to grasping is as a preprocessing step for grasp planning. An alternative proposed by Van der Merwe, et al.~\cite{van2020learning} and later by Jiang, et al.~\cite{jiang2021synergies} is to use a shape completion objective as a regularization term in the context of grasp learning. As will be discussed more in Section~\ref{sect:merwe2}, both these methods use an implicit representation where the model takes position as input and infers the grasp quality at that query point. However, these models also predict the signed distance of the query point to the object surface. Both papers report that this extra regularization term can improve grasp quality inference.

\subsubsection{Summary}

The idea of using shape completion methods to generate a complete geometry of an object from which grasps may be planned is attractive. Unfortunately, these methods have not yet been wildly successful. Shape completion is clearly a very challenging inference problem and it could be that direct grasp inference is simply easier.



\subsection{Implicit Shape Approaches to Grasp Inference}

Recently, spurred by the success of NeRF~\cite{mildenhall2020nerf} as an approach to novel view synthesis in computer vision, a few approaches have have explored implicit models for grasping. These methods take as input both a volumetric representation of the scene and a set of 3D grasp query positions and infer the grasp quality and gripper orientations associated with the input query points.


\subsubsection{Van der Merwe, et al (2020)}
\label{sect:merwe2}

One of the earliest works on implicit 6DOF grasp detection is the work of Van der Merwe, et al~\cite{van2020learning} which they dubbed \emph{PointSDF}. This model takes shape observations as input along with a set of 3D query points and infers the grasp quality and gripper orientation at the query points. The shape is input as a partial point cloud and passed to a PointConv model~\cite{wu2019pointconv}. This provides a feature representation of the global shape the is then flattened, combined with the 3D positions of the query points, passed through multiple fully connected layers, combined with the grasp orientations at the query points, and finally passed through more fully connected layers to produce a grasp quality prediction. Since grasp pose is an input to this model, a key challenge with this approach is to identify a good grasp pose. This is accomplished by sampling an initial pose and then following the gradient of the neural model calculated with respect to the grasp orientation input until reaching a maxima (this is similar to the refinement step of~\cite{mousavian20196}). A key aspect of this work is the addition of shape reconstruction as a regularization objective (see Section~\ref{sect:merwe1}). In addition to inferring grasp quality, this model also infers the signed distance from the object surface of the query points. The authors presented results that demonstrate that optimizing for this additional objective during training significantly improves the performance of grasp inference. Ultimately, the method was demonstrated to be able to achieve a 60\% grasp success rate on eight household YCB objects presented in isolation.


\subsubsection{Jiang, et al. (2021)}

Another approach that takes a volumetric TSDF representation as input is the Implicit Geometry and Affordance (GIGA) method of Jiang, et al.~\cite{jiang2021synergies}. This approach is similar to the method of Van der Merwe, et al.~\cite{van2020learning}: they represent the 3D scene using an implicit model and they regularize grasp predictions using an auxiliary shape reconstruction loss (in this case an occupancy probability rather than a signed distance.). Perhaps the most interesting thing here is the model itself: the model passes the TSDF input to a 3D convolutional layer. Then, the features are projected orthographically onto three orthogonal basis planes and passed to a standard UNet model. Finally, the UNet feature maps are projected to some fully connected layers with skip connections which output both grasp and occupancy predictions. A key difference here relative to Van der Merwe, et al. is that the model produces grasp orientation as an output rather than an input -- this makes it much easier to locate grasps. As in Van der Merwe, et al., these authors also report results that show that the auxiliary shape reconstruction objective during training can improve grasp detection performance. Ultimately, the model is demonstrated to achieve a 70\% grasp success rate using a Panda arm for densely cluttered household objects presented on a tabletop.


\subsection{Inferring $\SE(3)$ Grasps Using $\SE(2)$ Methods}
\label{sect:se3viase2}

Whereas most approaches to $\SE(3)$ grasp detection infer grasp pose from point cloud or voxel input, a few methods attempt to extend $\SE(2)$ methods.

\subsubsection{Berscheid et al. (2021)}
\label{sect:berscheid}

This approach uses a fully convolutional model to infer the position and top-down orientation of a grasp, and then use heuristics to determine the out-of-plane orientation variables~\cite{berscheid2021robot}. They infer the planar position and orientation of a grasp using a fully convolutional model comprised of a stack of eight dilated convolutional layers. As in~\cite{zeng2018learning}, they provide a stack of rotated input images to the model, thus implicitly evaluating grasp quality for a stack of different gripper orientations. Once an $x,y,\theta$ pose is selected using the fully convolutional model, the hand-coded strategy selects the two out-of-plane orientations that are used when executing the grasp. They frame grasping as a contextual bandit problem with Boltzman exploration. The result is an ability to grasp in 6-DOF that one would expect to learn at a similar rate to planar grasp methods. In bin picking experiments for challenging household objects presented in clutter, the authors report their method achieves a 92.1\% grasp success rate on the first 20 objects grasped from a 30 object bin.


\subsubsection{Kasaei and Kasaei (2021)}

This approach, dubbed \emph{MVGrasp}, also uses a fully convolutional 2D model to infer grasps in 3D~\cite{kasaei2021mvgrasp}. However, instead of determining out-of-plane orientations using heuristics, this method simply re-runs the 2D procedure from multiple different out-of-plane approach directions. The method starts by segmenting an object of interest. Then, an Eigenvalue analysis of the convariance matrix of the object points is performed, thereby producing three orthogonal approach directions and three corresponding planar views. One of these views is selected based on pixelwise image entropy and then grasp detection is performed using a fully convolutional model. The authors report that the method can achieve a 92\% grasp success rate for household objects in clutter. The results show their method performs favorably relative to GPD~\cite{tenpas_ijrr2017} and GG-CNN~\cite{morrison2018closing}.

\subsubsection{Summary}

Approaches to $\SE(3)$ grasping based on $\SE(2)$ methods are simple and work well in practice. However, these methods are not as flexible as other $\SE(3)$ grasp learning methods and will not suit all applications.


\subsection{Summary of $\SE(3)$ Methods}

Unlike $\SE(2)$ grasping where the Cornell benchmark has become a standard, there are no benchmarks that are widely used in $\SE(3)$ grasping. As a result, it is difficult to compare $\SE(3)$ grasping methods in a precise way. Here, we list grasp success rates from physical robotics experiments as shown in Table~\ref{table:se2_comparison}. 
Grasp success rates (GSR) are listed in one of two settings: either the \emph{Singulated Objects GSR} where objects are presented apart from others on a table or the \emph{Clutter Pile GSR} where objects are piled up randomly, as in the bin picking setting. The \emph{Object Type} column in Table~\ref{table:se3_comparison} characterizes objects in one of three ways. \underline{Household:} Common household objects resembling cylinders, boxes, and other amorphous object shapes; does not include screwdrivers or other thin objects. \underline{Challenging:} Complex or thin household object shapes such as screwdrivers, forks, or objects with complex geometry. \underline{Adversarial:} 3D printed objects of shapes considered to be challenging to grasp, e.g. the DEXNET adversarial objects~\cite{mahler2018guest} or the EGAD! objects~\cite{morrison2020egad}. The \emph{Camera View} column describes the number of RGBD cameras used. Most of these methods are ``single'' view, meaning that only a single camera is used. The six vertical groups in Table~\ref{table:se3_comparison} correspond to the methods covered in Sections~\ref{sect:hypandtestse3} through~\ref{sect:se3viase2}, respectively.

\begin{table}[h]
\tabcolsep7.5pt
\caption{Performance of the $\SE(3)$ grasp methods reviewed in this section.}
\label{table:se3_comparison}
\begin{center}
\begin{tabular}{@{}l|c|c|c|c|c@{}}
\hline
& Singulated  & Clutter & &  & \\
& Objects  & Pile & Object & Camera & \\
Method  & GSR$^{\rm a}$ & GSR$^{\rm b}$ & Type$^{\rm c}$ & View & Robot \\
\hline
Gualtieri, et al.~\cite{gualtieri2016high}   & & 93\%   & Household & Multi & Baxter \\
Liang et al.~\cite{liang2019pointnetgpd}   & & 89.33\%   & Household & Single & UR5 \\
Mousavian et al.~\cite{mousavian20196}   & 88.3\% &   & Household & Single & Panda \\
\hline
Qin et al.~\cite{qin2020s4g} & & 77.1\%   & Household & Single & Kinova \\
Sundermeyer et al.~\cite{sundermeyer2021contact}   &  & 84.31\% & Household & Single & Panda \\
Wu et al.~\cite{wu2020grasp}   & 85\% &   & Household & Single & UR5 \\
Zhao et al.~\cite{zhao2021regnet}   &  & 79.34\%  & Challenging & Single & Baxter \\
Wei et al.~\cite{wei2021gpr}   &  & 69.2\%  & Challenging & Single & ABB Yumi \\
\hline
Breyer et al.~\cite{breyer2021volumetric}   &  & 80\%  & Household & Multi & Panda \\
Cai et al.~\cite{cai2022real}   & &  78.43\% & Adversarial & Single & Barrett \\
\hline
Varley et al.~\cite{varley2017shape}   & 93\% &  & Household & Single & Barrett \\
Lundell et al.~\cite{lundell2019robust}   & 58\% &   & Household & Single & Panda \\
\hline
Van der Merwe, et al.~\cite{van2020learning}   & 60\% &   & Household & Single & Panda \\
Jiang, et al.~\cite{jiang2021synergies}   &  &  86.9\% & Household & Single & Panda \\
\hline
Berscheid et al.~\cite{berscheid2021robot}   &  & 92\%$^{\rm d}$  & Challenging & Single & Panda \\
Kasaei and Kasaei~\cite{kasaei2021mvgrasp}   &  & 92\%  & Household & Single & UR5 \\
\hline
\end{tabular}
\end{center}
\begin{tabnote}
$^{\rm a}$Grasp success rate for singulated objects on a table top.\\
$^{\rm b}$Grasp success rate for objects piled in clutter, as in the bin picking setting.\\
$^{\rm c}$Object type categorization is made by the author of this review.\\
$^{\rm d}$Grasp success rate for first 20 out of 30 objects grasped.
\end{tabnote}
\end{table}

\subsubsection{Obstacles to a Fair Comparison}

It is critical to recognize that the grasp success rates listed in Table~\ref{table:se3_comparison} are not directly comparable to each other. The problem is that different researchers use different benchmarks for their physical experiments. There are three main issues. \underline{Different object sets:} This is the biggest problem. The object set used to evaluate physical grasping performance is slightly different for nearly every method discussed in this section! One paper might have a lower grasp success rate because they evaluate on more challenging objects while another might have a high grasp success rate because they evaluate on easy objects. \underline{Different perceptual systems:} Most of the methods reviewed here evaluate physical performance using a point cloud generated by a single depth sensor. However, some methods like~\cite{gualtieri_iros2016} and~\cite{breyer2021volumetric} generate a fused point cloud or voxel grid using SLAM or a point registration technique. This gives those methods an advantage because they ``see'' more of the world than would be visible to a single sensor. \underline{Different Robots:} For the most part, this is a minor difference. With the exception of the Baxter, Kinova, and Barrett robots, these robots all have very precise position control and should therefore be comparable.

\subsubsection{Discussion}

The results shown in Table~\ref{table:se3_comparison} do not suggest any obvious conclusions. As noted above, one must not simply rank methods by their performance in the \emph{Clutter Pile GSR} column. That reasoning would suggest that Gualtieri et al.~\cite{gualtieri_iros2016,tenpas_ijrr2017} is the best performing method -- something that is certainly untrue because more than half the methods listed benchmark directly against~\cite{gualtieri_iros2016} and show outperformance. Nevertheless, a few methodological conclusions can be drawn. First, the hypothesise and test methods of Section~\ref{sect:hypandtestse3} seem a little dated. These methods are inefficient because they require a grasp classification model to evaluate a large set of candidates. Second, a variety of models have been explored based on  PointNet or PointNet++ architectures and none of these seems to be a clear winner. Third, the two methods based on shape completion, Varley et al.~\cite{varley2017shape} and Lundel et al.~\cite{lundell2019robust} seem to underperform the other methods significantly. For the other two major model architecture classes -- 3D convolutions and implicit models -- these could be promising directions for research, but it is too early to say. The two methods based on extending $\SE(2)$ archiectures into $\SE(3)$ -- Berscheid et al.~\cite{berscheid2021robot} and Kasaei and Kasei~\cite{kasaei2021mvgrasp} -- seem to perform very well in practice, but the fact that they are essentially $\SE(2)$ methods seems like a major limitation.




\section{GRASP METRICS AND EVALUATION}

Like many applications of machine learning, grasp learning methods are typically evaluated empirically. As such, performance metrics and evaluation datasets are essential when evaluating how well a new method works.

\subsection{Performance Metrics}

There is a surprising variety of performance metrics used in grasp learning. This variety reflects the different settings in which grasp methods operate and the performance criteria relevant to different applications.

\subsubsection{Grasp Success Rate}

The most basic grasp performance metric is the \emph{grasp success rate}, the proportion of grasp attempts that are successful, i.e. grasp trials where the robot has successfully grasped and lifted an object above a threshold height. This can be measured either in simulation or on a physical robotic system. Grasp success rate is often measured in two different settings: when grasping objects that are presented \emph{in isolation} (i.e. there is nothing else on the table) or when grasping objects presented \emph{in dense clutter}, i.e. when objects have been deposited into a pile or bin in an unstructured way. Grasping objects in isolation is generally considered to be an easier problem. Although grasp success rate is obviously an important metric, it does not capture the difficulty of the grasp setting and it ignores how the object was grasped.

\subsubsection{$n$ out of $m$ Grasp Success Rate}
\label{sect:noutofm}

When grasping in dense clutter all objects from a bin, it is typically the case that the algorithm attempts to grasp the easiest objects first, thereby obtaining higher grasp success rates early during bin clearing and lower grasp success rates later. As a result, it is common to measure grasp success rates separately for the first $n$ out of $m$ objects. For example, Kalashnikov et al. evaluate grasp performance using bins that contain 30 objects. They measure grasp success rates separately for the first 10, 20, and 30 objects~\cite{kalashnikov2018qt}.

\subsubsection{Completion Rate}

A high grasp success rate does not imply that a system is able to reliably remove all items from a bin -- it could be that the system always fails to grasp the last item. Nevertheless, bin clearing is often an important objective in industry. The ability of a system to clear a bin is captured by \emph{completion rate}. It is assumed that a series of bin clearing trials are performed where the bins are always loaded with the same number of objects. The \emph{completion rate} is the proportion of bin clearing trials that were successful, i.e. where the robot cleared all objects from the bin~\cite{wei2021gpr}. Grasp completion rate and grasp success rate are in tension in the sense that lowering the confidence threshold at which a system is willing to attempt a grasp could increase completion. As a result, it is possible to carve out a pareto front, similar to a precision-recall curve, by varying the grasp acceptance threshold. This success-completion curve itself is another relevant grasp performance metric.


\subsubsection{Coverage Rate}

Another metric relevant to grasping is \emph{coverage}, the degree to which a grasp algorithm can find all the ways in which an object can potentially be grasped~\cite{mousavian20196}. Given a set of grasps that are known to exist on an object, the \emph{coverage rate} is the proportion of those grasps that are detected by the algorithm. This measure is most relevant to $\SE(3)$ grasping methods rather than $\SE(2)$ methods because it is only in $\SE(3)$ that there are enough different ways of grasping an object to be concerned about coverage. Together with the grasp success rate, the coverage rate also define a pareto front similar to the success-completion curve described above or the precision-recall curve. In fact, the coverage rate is exactly analogous to recall (the proportion of true positives that are detected by a classifier). Unfortunately, the notion of coverage in grasping suffers from a key problem -- the space of feasible grasps on an object is continuous, not discrete. As a result, in order to define coverage, we must \emph{sample} the space of feasible grasps in order to create a grasp set. However, since the coverage rate will be slightly different depending upon how grasps are sampled, it is difficult to obtain a fair comparison using this metric unless the set of possible feasible grasp samples is fixed per object -- something which is possible, but not currently part of most object datasets.

\subsubsection{Picks Per Hour (PPH)}

Another grasp metric that has been used in the literature is \emph{Picks Per Hour} (PPH)~\cite{satish2019policy}. This metric simply measures the number of objects that are grasped successfully by a physical robotic system in the space of an hour. This metric is nice because it measures throughput -- something which is often important in industrial applications. Unfortunately, because it measures end-to-end performance, PPH depends a lot on the implementation details of the system (e.g. how fast the robot arm can move) and may not reflect the potential of the method.

\subsubsection{Intersection Over Union (IOU)}
\label{sect:iou}

\emph{Intersection Over Union} (IOU) is a metric that is predominantly used in object detection, but was adopted into grasp detection by~\cite{jiang_icra2011,lenz_rss2013}. IOU is typically used in the context of measuring performance using $\SE(2)$ datasets such as the Cornell Dataset~\cite{jiang2011efficient} and the Jacquard Dataset~\cite{depierre2018jacquard}. Since these are $\SE(2)$ datasets, a grasp can be expressed as a box in the image that is centered and aligned with the gripper position and orientation. If desired, the length of the box can encode gripper width. Given a ground truth bounding box showing the true pose of a grasp and an inferred bounding box showing the predicted pose of a grasp, the IOU measures the ratio between the intersection of these boxes and the union of the boxes. If the boxes are disjoint, then IOU is zero. Using IOU, it is possible to evaluate grasp performance as detection accuracy. Given a set of ground truth grasps known to exist in an image, accuracy is the proportion of these grasps for which the IOU (measured with respect to the nearest grasp prediction) is above some threshold, often 25\% or 30\%. Often, the orientation error between the grasps must also be below a threshold, typically 30 degrees. Grasp performance measured using Cornell or Jaquard is nearly universally measured using this metric.

\subsubsection{Instance-wise Versus Object-wise Splits}

The objective of grasp learning is typically to be able to grasp novel objects -- objects not seen during training. This requires us to be careful about how the dataset is ``split'' between training and testing. There are two obvious choices regarding how to split the data: the instance-wise split and the object-wise split. With the \emph{instance-wise split}, the training and test sets may contain objects from the same object categories, e.g. there might be coffee mugs in both the train and test sets. However, the test set may only contain object instances (i.e. specific object geometries) not seen during training. With the \emph{category-wise split}, we go one step further and require the test set to contain objects belonging to categories not seen during training. So, for example, if there are coffee mugs in the test set, then there may not be any coffee mugs in the training set. Obviously, the object-wise split is the more challenging scenario and we would expect grasp success rates to be lower there.



\subsection{Grasp Datasets in $\SE(2)$}

Several datasets are available for measuring grasp performance in $\SE(2)$. These datasets are typically comprised of a set of images, each of which is labeled with one or more ground truth grasps.

\subsubsection{Cornell Dataset (2011)}
\label{sect:cornell}

The first grasp dataset to have a major impact on robotic grasp learning was the Cornell Dataset~\cite{jiang2011efficient}. This dataset is comprised of 885 top-down RGBD images paired with a total of 8019 ground truth grasps that are hand labeled. As described in Section~\ref{sect:iou}, a grasp is expressed as a grasp rectangle defined by five parameters: two parameters that denote rectangle position, two parameters that denote gripper height and width, and one parameter that denotes orientation. Grasp detection performance is measured as detection accuracy with respect to the IOU metric, generally with a 25\% IOU threshold and a 30\% orientation threshold.



\subsubsection{Jacquard Dataset (2018)}

The Jacquard dataset can be viewed as a larger and more objective version of the Cornell Dataset. Jacquard contains 54K images generated in simulation using Blender. The images are generated using 11K unique object mesh models. Each image is labeled with multiple grasp rectangles (the same representation used in Cornell) that roughly cover the space of possible grasps, for a total of 1.1M grasp labels. The labels are generated automatically by sampling a large number of possible grasps at random and simulating them in PyBullet to evaluate their quality. As with Cornell, performance in Jacquard is measured as accuracy with respect to a 25\% or 30\% IOU detection threshold.



\subsubsection{DexNet Dataset (2017)}

Another planar dataset that is sometimes used is the DexNet 2.0 dataset comprised of approximately 6.7M depth images created using 1500 mesh models generated in simulation~\cite{mahler2017dex}. Each depth image is labeled with up to 100 grasp detections, obtained using a grasp planning method on the underlying mesh models. These object models were selected from 3DNet and the KIT Object Database and are drawn from 50 different object categories. The dataset is augmented with per-pixel Gaussian noise and by image reflections and rotations of 180 degrees.

\subsection{Grasp Datasets in $\SE(3)$}

There are also several grasp and object datasets available for use in $\SE(3)$ grasping, a few of which we highlight below.


\subsubsection{Columbia Grasp Database (2009)}

This is probably the earliest grasp dataset for $\SE(3)$ grasping~\cite{goldfeder2009columbia}. It is comprised of 7256 object mesh models drawn from the Princeton Shape Benchmark~\cite{shilane2004princeton}. In total, 238k $\SE(3)$ grasps are labeled over the 7256 models using the GRASPIT! simulator~\cite{miller2005robotic}.

\subsubsection{YCB (2015)}

YCB (Yale, Columbia, Berkely)~\cite{calli2015benchmarking} is an object set rather than a grasp dataset, but it is used so frequently that we include it here. YCB is a set of 77 household objects for which 3D mesh models and RGB images are available online. Perhaps the most notable thing about YCB is that there was an effort in 2015 and 2016 to distribute physical versions of this object set to members of the research community working in the field.

\subsubsection{Dex-Net Objects, Including Adversarial Objects (2017)}

The Berkeley Dex-Net project was published with the following three objects sets~\cite{dexnetwebsite,mahler2017dex}. 1) 1.3k synthetic 3D mesh models from a 50 category subset of 3DNet; 2) 129 mesh models from the KIT Object Database; 3) 13 ``adversarial'' models. Of these, the 13 adversarial objects is arguably the most important because researchers can easily print these objects and benchmark their own methods on a very challenging object set.

\subsubsection{EGAD! (2020)}

EGAD! is an interesting dataset comprised of 2331 objects~\cite{morrison2020egad}. The focus here is on an algorithmic approach to generating 3D object geometries that span a space of shape complexity and grasp difficulty. In particular, each object is ranked from 1 to 25 according to both measures (shape complexity and grasp difficulty) and placed on a $25 \times 25$ grid. The 2331 objects were generated while ensuring that the dataset contains no more than four objects belonging to any single grid square. The result is a quantiatively diverse dataset. Out of this dataset, 49 object meshes are designated as evaluation objects, to be used in benchmarking grasp algorithms. While the main focus is on the object meshes, the dataset includes a grasp training set created using the grasp planning and labeling method of~\cite{mahler2017dex}.


\subsubsection{GraspNet-1Billion (2020)}

GraspNet-1Billion is another grasp dataset the bears mentioning~\cite{fang2020graspnet}. Although it includes data drawn from only 88 objects (32 of which are from YCB and another 13 from DexNet 2.0), it is paired with 97k of RGBD images of 190 different cluttered scenes containing the objects. Each of these RGBD images is labeled with tens of thousands of $\SE(3)$ grasp poses, for a total of 1.1B labeled grasp examples. In addition to the grasps, each scene is labeled with the $\SE(3)$ poses of the objects as well.


\subsubsection{ACRONYM (2021)}

ACRONYM is a recent $\SE(3)$ dataset comprised of 17.7 million labeled grasps performed on 8872 objects meshes~\cite{eppner2021acronym}. The object meshes were obtained from ShapeNet~\cite{savva2015semantically}. The grasps were labeled using the FleX physics simulator~\cite{macklin2014unified} to simulate the Franka Panda gripper (maximum grip aperture of 8cm). Observations are available either as depth images or point clouds, produced using PyRender2. A notable aspect of this dataset is that it does not include data in the standard dense clutter setting. Instead, it includes two types of scenes: 1) a single object setting where a single object is presented floating in space; 2) a ``structured clutter'' setting where one or more objects have been placed on top of a surface in a way that is typical in domestic applications.




\section{Discussion}

\subsection{Grasp Success Rates}

Perhaps the most important and surprising observation to make based on the \emph{Clutter Pile GSR} columns in Tables~\ref{table:se2_comparison} and~\ref{table:se3_comparison} is that grasp success rates for challenging novel household objects in clutter are still somewhere in the mid 80\% to low $90\%$ range. In other words, most bin picking systems for novel objects in clutter will fail once out of every ten grasp attempts. It is not clear why this is the case. One answer is that we are not yet using large or complex enough models trained on sufficiently large datasets. Another possibility is that we are suffering from the sim2real gap. Since most grasp systems are trained in simulation, it could be that we lose 10\% grasp performance on the real system. A third possibility could be that we are not bringing enough structure to the problem, e.g. perhaps object factored models would perform better. In any case, it is also unclear whether a 90\% grasp success rate is a problem. In many bin picking settings, grasp failure is not catastrophic because it is usually possible simply to try again. As Satish et al.~\cite{satish2019policy} point out, the more relevant metric in bin picking applications is probably mean picks per hour (MPPH). Another important point is that some industrial applications require grasping \emph{known} objects, not \emph{novel} objects. In these applications, it should be possible to reach grasp success rates near 100\%. Nevertheless, there are clearly settings where high grasp success rates for novel objects are important and further research is clearly needed to achieve this.

\subsection{Benchmarking}

Another point to make is that the field is obviously suffering from a lack of widely accepted benchmarks. The exception is the Cornell Detection benchmark which is widely used in $\SE(2)$ grasping. Unfortunately, there is nothing comparable for $\SE(3)$ grasping and this makes the comparison between $\SE(3)$ methods very difficult. One candidate for a ``3D version'' of Cornell is the ACRONYM dataset~\cite{eppner2021acronym}, a dataset that designates a large set of training and test grasps in $\SE(3)$. Other possibilities are EGAD! which provides an interface for obtaining labeled grasps on both training and test objects or GraspNet-1Billion~\cite{fang2020graspnet} which provides a set of RGBD images labeled with 3D grasps. However, even if the field does not standardize on a dataset, even something simple like standardizing on a single set of evaluation objects for physical robotic grasping experiments would help a lot. One good candidate for this is to 3D-print the 49 EGAD! evaluation objects designated as such in the EGAD! object set~\cite{morrison2020egad} and to use them to evaluate in a bin picking scenario. An alternative that requires less 3D-printing is to follow the same procedure with the 13 Adversarial Dex-Net objects. Finally, benchmarking using the YCB object set~\cite{calli2015benchmarking}, comprised of commercially available consumer items, is another possibility.


\subsection{Outlook}

Since it is difficult to make accurate predictions about future technical developments, I will limit myself to making just a few comments. First, this is an exciting time for grasp learning research. Today, it has become commonplace to see robotic grasping of novel objects from cluttered bins. This is something that was simply not possible ten years ago. Recent developments in deep learning should bear much of the credit for this success. Grasp learning research now appears to be closely tied to trends in machine learning and computer vision and this seems like something that is likely to continue. However, it would be a mistake to view grasp learning as simply an application of the same machine learning methods that have been developed in computer vision, NLP, and elsewhere. There are many aspects of robotics that make it a unique and interesting application domain. First, unlike other applications of machine learning which generally focus on regression and classification, robotics problems fundamentally involve policy learning. This necessitates a focus on reinforcement learning, learning from demonstration, and other forms of policy learning. Second, robotics problems generally have a geometric structure that is often not present in NLP or even computer vision. This is something that is likely to receive more attention in the future.

\bibliographystyle{ar-style3.bst}

\bibliography{references}

\end{document}